\documentclass{article}
\usepackage{spconf,amsmath,graphicx}
\usepackage{booktabs}
\usepackage{multirow}
\usepackage{amsmath}
\usepackage{amssymb}
\usepackage{caption}
\usepackage{subcaption}
\usepackage{cite}

\title{In-Bed Human Pose Estimation from Unseen and \\ Privacy-Preserving Image Domains}
\name{Ting Cao$^{1,3}$, Mohammad Ali Armin$^{3}$, Simon Denman$^{2}$, Lars Petersson$^{3}$, David Ahmedt-Aristizabal$^{2,3}$ }
\address{
$^{1}$ Australian National University, Canberra, Australia. \\
$^{2}$ SAIVT, Queensland University of Technology, Brisbane, Australia. \\
$^{3}$ Imaging and Computer Vision Group, CSIRO Data61, Canberra, Australia. \\
\tt\normalsize david.ahmedtaristizabal@data61.csiro.au
}

\begin{document}
\ninept
\maketitle

\begin{abstract}
Medical applications have benefited greatly from the rapid advancement in computer vision. Considering patient monitoring in particular, in-bed human posture estimation offers important health-related metrics with potential value in medical condition assessments.
Despite great progress in this domain, it remains challenging due to substantial ambiguity during occlusions, and the lack of large corpora of manually labeled data for model training, particularly with domains such as thermal infrared imaging which are privacy-preserving, and thus of great interest.
Motivated by the effectiveness of self-supervised methods in learning features directly from data, we propose a multi-modal conditional variational autoencoder (MC-VAE) capable of reconstructing features from missing modalities seen during training. This approach is used with HRNet to enable single modality inference for in-bed pose estimation.
Through extensive evaluations, we demonstrate that body positions can be effectively recognized from the available modality, achieving on par results with baseline models that are highly dependent on having access to multiple modes at inference time.
The proposed framework supports future research towards self-supervised learning that generates a robust model from a single source, and expects it to generalize over many unknown distributions in clinical environments.
\end{abstract}

\vspace{-5pt}
\section{Introduction}
\label{sec:intro}
\vspace{-5pt}

Reliable patient posture estimation provides quantifiable data that can offer an objective assessment of a patient's behavior. Such movement assessment is a powerful tool during clinical observations to aid the diagnosis of motor and mental disorders, as well as sleep disorders such as apnea. 
Vision-based systems using deep learning, with their non-invasive nature, have shown potential to extract comprehensive pose through granular quantification of a patient's position~\cite{ahmedt2019understanding,hesse2019learning,ahmedt2018deep}. 
Yet, these methods are hindered by hospital conditions including having the patient's body covered; and the requirement of richly labeled visible data (RGB) with in-bed human poses.
While RGB data can be used to recognize the actions and position of a subject, its use is limited due to privacy concerns.
Thus, in order to see through covers and to provide stable image quality under lighting variation, only privacy preserving modalities such as Long-Wave Infrared (LWIR) imaging are used to monitor patient activities. The additional domain gap between RGB and LWIR imaging raises great challenges for pose estimation methods.

\begin{figure}[!t]
\centering
\includegraphics[width=1\linewidth]{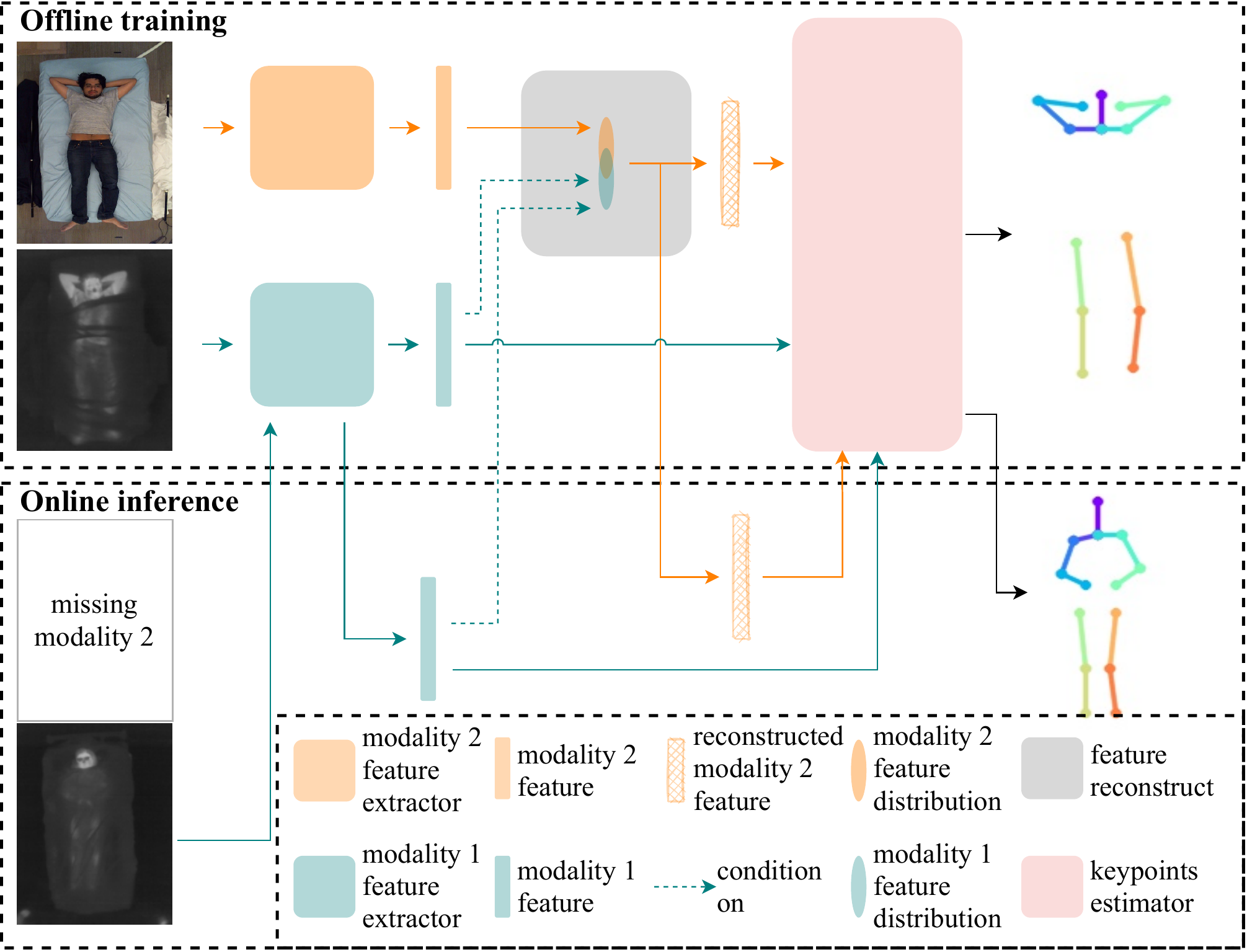}
\vspace{-12pt}
\caption{
Overview of the proposed self-supervised framework. 
\textit{Offline training}: a feature reconstruction module is adopted to learn the distribution of a modality which is absent at inference time through a variational autoencoder. During training we have access to both modalities, RGB and LWIR in the current setup.
\textit{Online inference}: the system only requires information from the available modality (LWIR) as the features of the missing modality can be reconstructed.
}
\label{fig:Fig1}
\vspace{-15pt}
\end{figure}

\begin{figure*}[!t]
     \centering
     \begin{subfigure}[b]{0.33\textwidth}
         \centering
         \includegraphics[width=\textwidth]{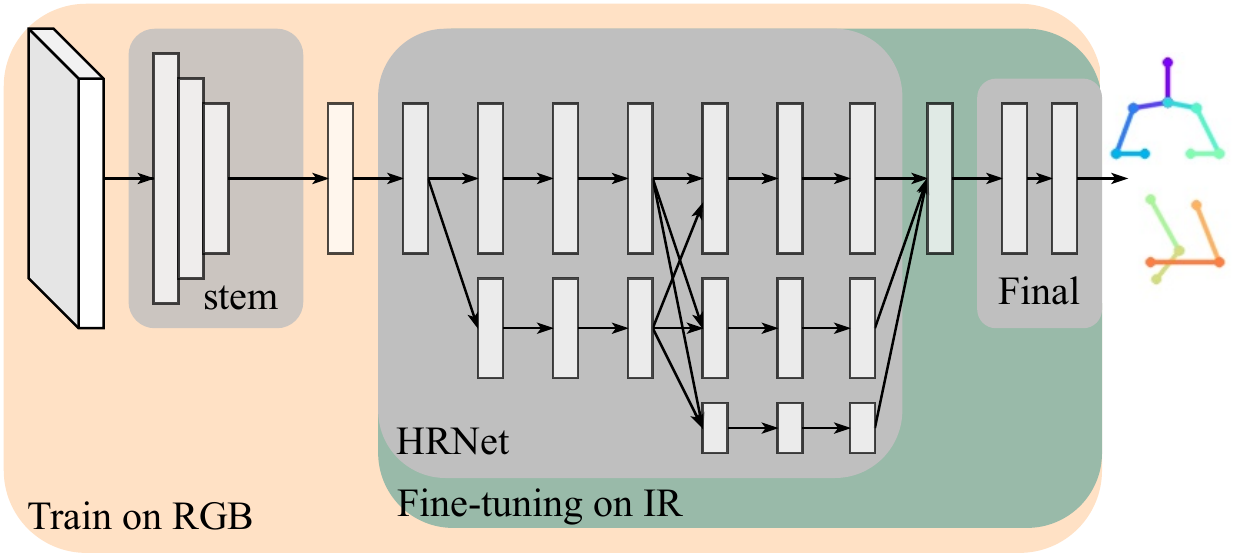}
         \caption{Fine-tuning}
         \label{fig:fine-tuning}
     \end{subfigure}
     \begin{subfigure}[b]{0.34\textwidth}
         \centering
         \includegraphics[width=\textwidth]{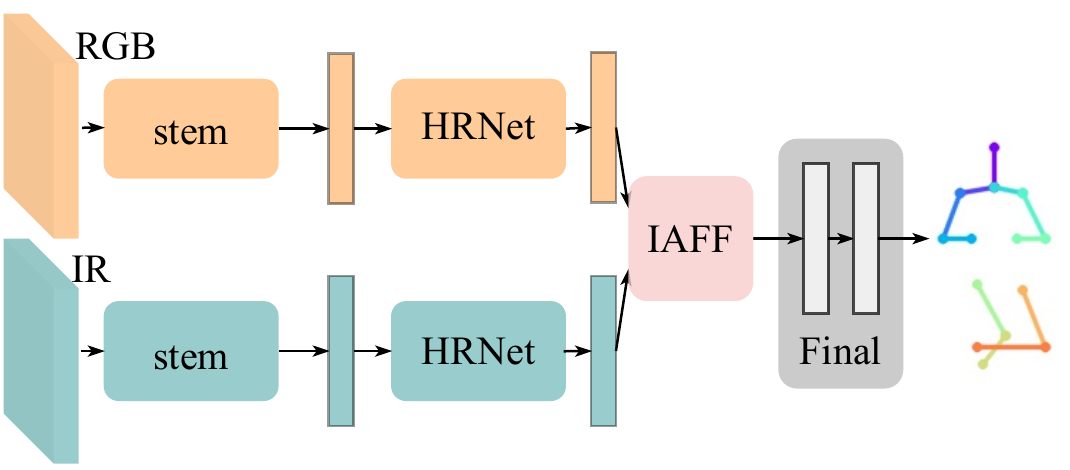}
         \caption{Feature fusion}
         \label{fig:feature-fusion}
     \end{subfigure}
     \begin{subfigure}[b]{0.33\textwidth}
         \centering
         \includegraphics[width=\textwidth]{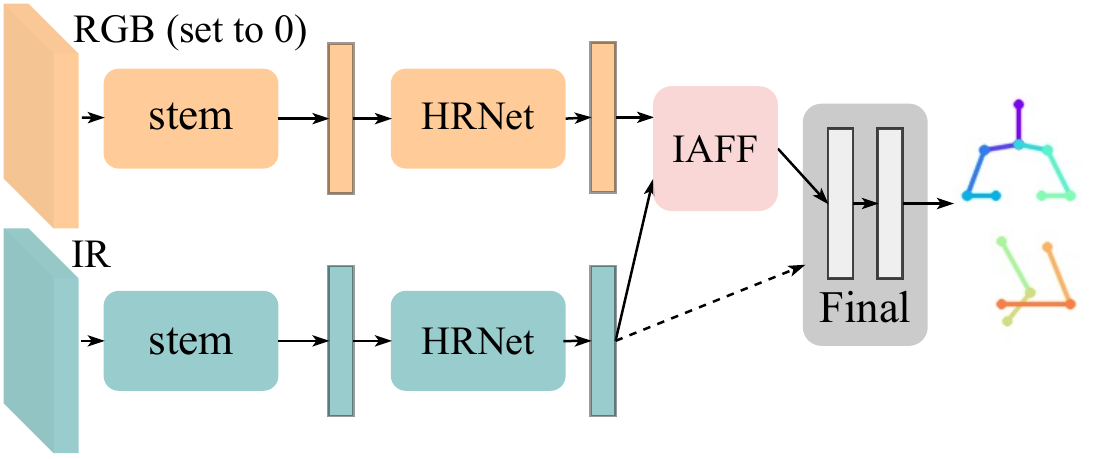}
         \caption{RDF}
         \label{fig:rdf}
     \end{subfigure}
    \begin{subfigure}[b]{0.40\textwidth}
         \centering
         \includegraphics[width=\textwidth]{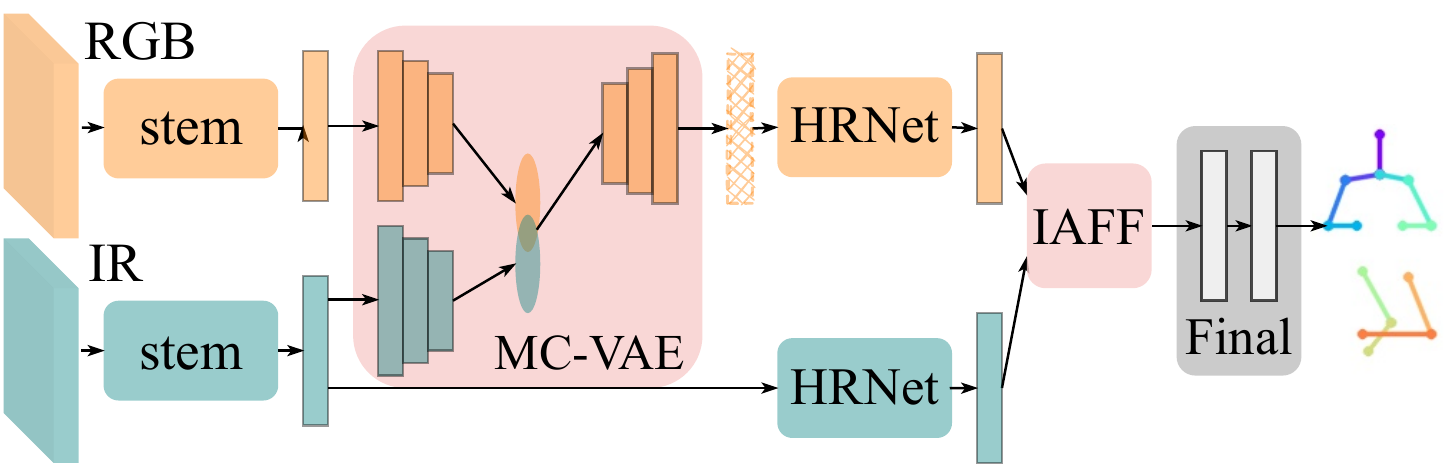}
         \caption{MC-VAE}
         \label{fig:ours}
     \end{subfigure}
        \vspace{-4pt}
        \caption{Comparison between baseline models and our self-supervised approach. 
        Models  (\protect\subref{fig:fine-tuning}), (\protect\subref{fig:feature-fusion}) and (\protect\subref{fig:rdf}) are based on existing methods. 
        (\protect\subref{fig:fine-tuning}) is trained on the secondary modality (RGB) then we froze the stem layer, fine-tuning the HRNet and final decision layer on the first modality (LWIR).
        (\protect\subref{fig:feature-fusion}) images from both modalities are required as input, a late-layer fusion module based on IAFF \cite{dai2021attentional} fuses the late features from parallel HRNet model.
        (\protect\subref{fig:rdf}) To simulate the missing modality scenario at inference time, RDF \cite{karanam2020towards} randomly sets the RGB image to 0 during training, during inference the RGB image is always 0 to represent the missing modality. When the RGB image input is 0, the model skips the fusion and uses features from IR streams directly (dotted line).
        (\protect\subref{fig:ours}) Our model leverages features from both modalities during training and reconstructs the missing modality conditioned on the available modality during inference.}
        \label{fig:three graphs}
\vspace{-8pt}
\end{figure*}

Prior work on for in-bed human pose estimation from LWIR domains has primarily relied on extracting features using transfer learning (fine-tuning), where the representation from networks trained on large-scale datasets (\textit{e.g.} COCO~\cite{lin2014microsoft}) are transferred to a small sample dataset~\cite{wang2021video,liu2019bed,liu2019seeing,chen2018patient,tajbakhsh2016convolutional}.
However, when fine-tuning for example an RGB model on thermal images, the low-level features learned on RGB images do not necessarily align with low-level thermal images features.

Leveraging multi-modal image sources via multi-modal fusion has significantly improved analysis results in various medical applications~\cite{dolz2018hyperdense, nie2016fully}, as well as in-bed pose analysis~\cite{yin2020multimodal, transue2017thermal}. 
These methods fuse features from different modalities to impose pose estimation accuracy. However, existing multi-modal in-bed pose analysis methods fail when one modality is unavailable during inference, as these models expect all training modalities as input. As such, these methods are unfeasible due to privacy issues raised in real-world medical monitoring scenarios.
%
Methods such as robust dynamic fusion (RDF)~\cite{karanam2020towards} can address the missing modality issue via a conditional training scheme. During training, an input modality is set to $0$ with probability $p$ to mimic the missing modality, and during inference the missing modality is also set to $0$. While this approach can address the missing modality issue, it does not explicitly learn or use features from missing modalities, and we argue that its bypass training scheme will essentially learn a separate model for situations where certain modalities are missing. Hence, it fails to truly leverage the multi-modal image sources.

Self-supervised methods are popularity to improve the quality of learned embeddings~\cite{kingma2013auto} in place of transfer learning. 
In this paper, we propose a novel \textit{multi-modal conditional variational autoencoder (MC-VAE)} for in-bed pose estimation.
Our approach is able to reconstruct the features of a missing modality with guidance from the features of the available modality, and thus addresses the issue of a missing modality. During training, the feature distribution of the visible modality is learned through a feature reconstruction module, and during inference when the mode is absent it can be generated from the LWIR modality. 
%
This architecture is shown in Fig.~\ref{fig:Fig1}.
This approach avoids privacy concerns that arise in low resource clinical environments (\textit{e.g.} just one sensor is needed), while explicitly leveraging reconstructed features to yield an accurate pose estimation. 
The contributions of our work are summarized as follows:
\begin{enumerate}
\vspace{-3pt}
\item We propose a robust method based on variational autoencoders for learning cross-modality representations, when deploying AI-based methods for patient monitoring where only one modality is present.
\vspace{-3pt}
\item We introduce a novel system to accurately estimate in-bed pose by reconstructing RGB features while preserving patient privacy within hospital environments.
\end{enumerate}

\begin{table}[!t]
\caption{Functionality comparison among methods. 
Feature fusion can fully learn from both modalities, but requires both modalities at inference to work. 
Both fine-tuning and RDF~\cite{karanam2020towards} can infer from a single modality (\textit{i.e.} when one modality is missing), but they can only partially learn from the secondary modality.
Our model fulfills both system requirements (SyR) by reconstructing features from secondary modality.}
\vspace{-5pt}
\centering
\resizebox{0.31\textwidth}{!}{%
\label{table:function}
\begin{tabular}{l c c}
\toprule
Approach & SyR1 & SyR2 \\
\midrule
Fine-tuning         & Yes       & Partial \\
Feature fusion      & No        & Yes  \\
RDF                 & Yes       & Partial  \\
(MC-VAE) (ours)         & Yes       & Yes \\ 
\bottomrule
\multicolumn{3}{p{150pt}}
{
SyR1: Inference on single modality. \newline
SyR2: Learn from secondary modality. \newline
} 
\end{tabular}}
\vspace{-24pt}
\end{table}

\begin{figure*}[!t]
\centering
\begin{subfigure}[b]{0.4\textwidth}
         \centering
         \includegraphics[width=\textwidth]{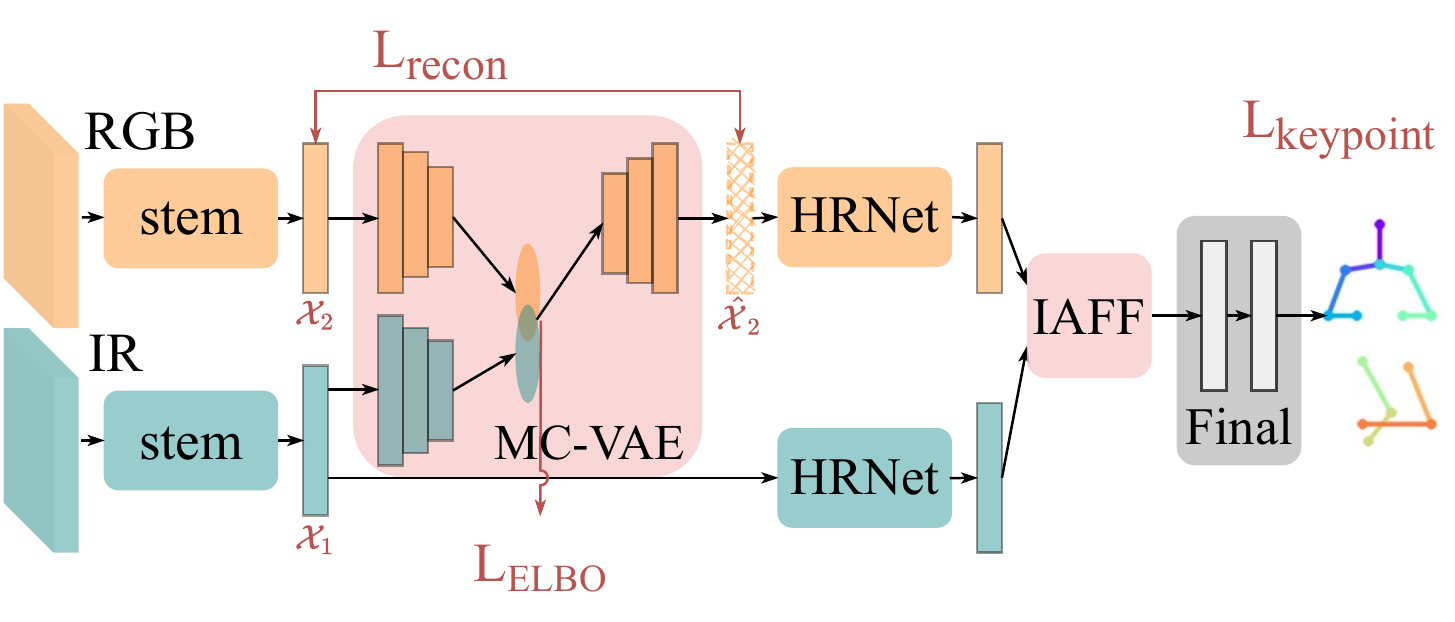}
         \caption{Training scheme}
         \label{fig:training}
     \end{subfigure}
     \begin{subfigure}[b]{0.43\textwidth}
         \centering
         \includegraphics[width=\textwidth]{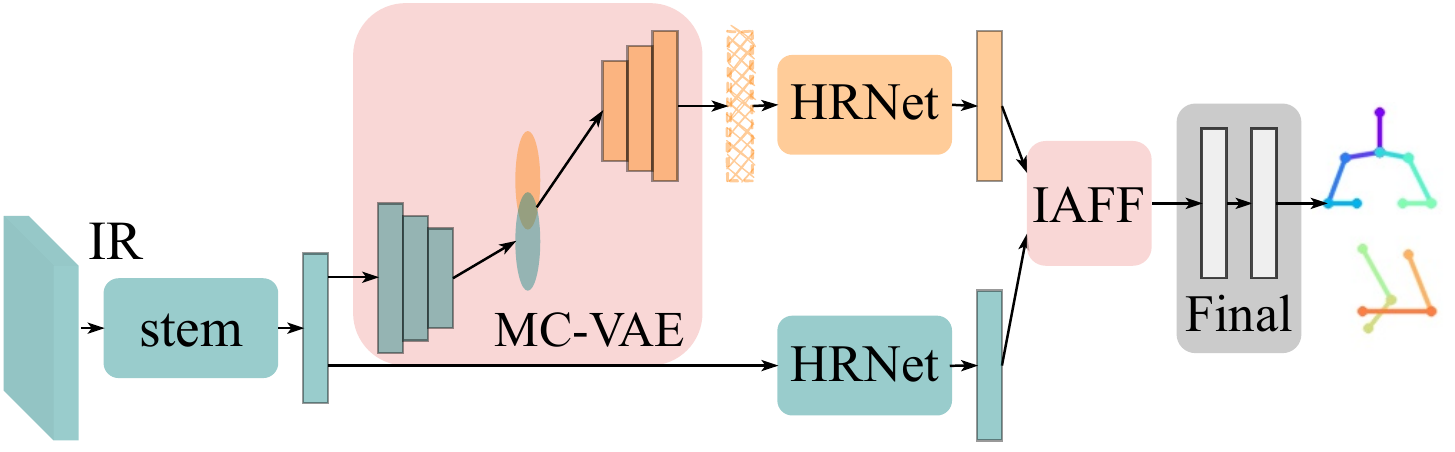}
         \caption{Inference}
         \label{fig:inference}
     \end{subfigure}
\vspace{-6pt}
   \caption{An illustration of the training and testing scheme. 
   (a) During training, both modalities are available, the model first extracts features $\mathcal{X}_1$ and $\mathcal{X}_2$ from each modality (shown in pink). Then the conditional variational autoencoder learns the joint representation of the  features from both modalities, and reconstructs feature $\hat{\mathcal{X}_2}$ conditioned on $\mathcal{X}_1$. The reconstructed feature $\hat{\mathcal{X}_2}$ is later trained in parallel with $\mathcal{X}_1$, then fused together via the Fusion module. The final fused features are used to predict joint locations.
   (b) During testing, modality 2 is unavailable, but the conditional variational autoencoder can reconstruct $\hat{\mathcal{X}_2}$ conditioned on $\mathcal{X}_1$ without input $\mathcal{X}_2$.
   }
\label{fig:train}
\vspace{-8pt}
\end{figure*}

\vspace{-12pt}
\section{Method}
\vspace{-5pt}

We compare three baseline architectures with our novel self-supervised approach that enables in-bed pose estimation when only privacy-preserving images (\textit{e.g.} infra-red images) are presented. Our approach can be extended to estimate poses in other unseen image domains as well. We compare the functionality of these four approaches in Table~\ref{table:function}, and empirically demonstrate the advantages of our proposed architecture. These architectures are illustrated in Fig.~\ref{fig:three graphs}.
For fair comparison, we (i) adopt HRNet \cite{sun2019deep} as our universal backbone across all methods, since it is currently the SOTA model in terms of efficiency and accuracy for pose estimation. (ii) employ iterative attentional feature fusion (IAFF) \cite{dai2021attentional} as our feature fusion method due to its superior performance and interpretability.

\vspace{-5pt}
\subsection{Existing multi-modal in-bed pose analysis methods}

Multi-modal in-bed pose analysis methods can be grouped according to the availability of the secondary modality during inference. 

Fine-tuning is employed as one of our baselines to address the issue of the absence of the secondary mode (Fig.~\ref{fig:fine-tuning}). We  first train our model using the RGB modality, then fine-tune the trained model on the LWIR modality.
%
In scenarios where the secondary modality is available during inference, existing works \cite{yin2020multimodal, transue2017thermal, clever2020bodies} utilize features from the secondary modality via feature fusion on intermediate layers. 
We adapt this design to create a baseline model illustrated in Fig.~\ref{fig:feature-fusion}. The model consists of two parallel HRNets, $\mathbf{H}_1$ and $\mathbf{H}_2$. It always requires input images, $\mathbf{I}_1$ and $\mathbf{I}_2$, from both modalities and then extract features $\mathbf{f}_1 = \mathbf{H}_1(\mathbf{I}_1)$ and $\mathbf{f}_2 = \mathbf{H}_2(\mathbf{I}_2)$ respectively. The fusion module fuses these two features and feeds the result to the final decision layers.

Inspired by \cite{karanam2020towards}, we employ the robust dynamic fusion (RDF) strategy as our third baseline (Fig.~\ref{fig:rdf}), where we set input image $\mathbf{I}_{2}$ to a tensor of zeros with probability $p$ to mimic the missing modality scenario during inference. 
When an input image $\mathbf{I}_{2}$ is a zero tensor, the model skips the fusion and directly takes feature $\mathbf{f}_1$ as the final feature. Therefore, when a modality becomes unavailable, the model can still perform inference with the learned parameters in $\mathbf{H}_1$.

\vspace{-5pt}
\subsection{In-bed pose estimation on unseen image domains}
Due to the nature of the hospital monitoring environment, RGB images cannot be used during inference as they violate hospital privacy protocols. Even if RGB images of patients were allowed, many inputs would be of limited value since patient joints are frequently heavily occluded by bedding.
To tackle these issues, we propose a novel approach to handle unseen modalities, where we explicitly reconstruct features for the missing modality. In this paper, for simplicity, we only consider a two modalities scenario (M=2). Training and inference schemes are presented in Fig.~\ref{fig:train}. 

We adopt a feature reconstruction module from~\cite{doersch2016tutorial, shi2019variational} to explicitly learn the distribution of the missing modality. As illustrated in Fig.~\ref{fig:train}, during training, multi-modal features $\mathcal{X}_1$ and $\mathcal{X}_2$ extracted by a backbone are the inputs to our multi-modal conditional variational autoencoder (MC-VAE). The joint variational posterior $q_\Phi(\mathbf{z}\;|\;\mathcal{X}_{1:M})$ can be factorized as
\begin{equation}
    q_\Phi(\mathbf{z}\,|\,\mathcal{X}_{1:2}) =
    \alpha_1\cdot q_{\phi_1}(\mathbf{z}\,|\,\mathcal{X}_1) + \alpha_2\cdot q_{\phi_2}(\mathbf{z}\,|\,\mathcal{X}_2) ,
\end{equation}
where $\alpha_1 = \alpha_2 = 1/2$, with the likelihoods $ q_{\phi_m}(z\;|\;\mathcal{X}_m)$ parametrized by encoders with parameters $\Phi = \{\phi_1, \phi_2\}$. We use the objective
\vspace{-3pt}
\begin{equation}
\vspace{-3pt}
    \mathcal{L}_{reg}=\mathcal{L}_{ELBO}(\mathcal{X}_{1:2}) =
    \mathbb{E}_{\mathbf{z} \sim q_{\phi}(\mathbf{z}\,|\,\mathcal{X}_{1:2})}\left[\log \frac{p_\Theta(\mathbf{z}, \mathcal{X}_{1:2})}{q_\Phi(\mathbf{z}\,|\,\mathcal{X}_{1:2})}\right] ,
\end{equation}
to approximate the true joint posterior $p_\Theta(\mathbf{z}_m^k, \mathcal{X}_{1:2})$. Once the joint posterior is learned, our feature reconstructor can rebuild the missing data using the available mode, \textit{i.e.} when $\mathcal{X}_2$ becomes unseen during inference, we can reconstruct the cross-modal feature $\hat{\mathcal{X}}_2$ conditioned on $\mathcal{X}_1$ via
\begin{equation}
    p(\hat{\mathcal{X}_2}\,|\,\mathcal{X}_1) = \mathbb{E}_{q(\mathbf{z}\,|\,\mathcal{X}_1)}\left[p(\hat{\mathcal{X}}_2\,|\,\mathbf{z})\right].
\end{equation}
To ensure that the reconstructed feature $\hat{\mathcal{X}}$ is suitable for later layers, we use the reconstruction loss from \cite{doersch2016tutorial}. We constrain $\hat{\mathcal{X}}$ to be between 0 and 1 using a Sigmoid function as per \cite{doersch2016tutorial}, thus $\hat{\mathcal{X}} \approx p(\hat{\mathcal{X}_1}\,|\,\mathcal{X}_2)$. Therefore, the reconstruction loss is given by
\begin{equation}
    \mathcal{L}_{Recon}(\hat{\mathcal{X}_2}, \mathcal{X}_2) =
    \left|\left|\mathbb{E}_{q(\mathbf{z}\,|\,\mathcal{X}_2)}\left[p(\hat{\mathcal{X}}_1\,|\,\mathbf{z})\right] - \mathcal{X}_2\right|\right|_2.
\end{equation}

By optimizing the above two objectives, we can obtain a feature reconstruction module which can take the presented feature $\mathcal{X}_1$ as input, and sample an authentic copy of the missing feature $\mathcal{X}_2$, $\hat{\mathcal{X}_2}$, conditioned on $\mathcal{X}_1$. 
This reconstruction capability enables inference on a single privacy preserving modality. As illustrated in Fig.~\ref{fig:Fig1}, during offline training we train our model using both modalities (LWIR and RGB), while simultaneously the feature reconstruction module learns the distribution of feature $\mathcal{X}_2$, enabling it to reconstruct feature $\hat{\mathcal{X}_2}$ on the condition of feature $\mathcal{X}_1$. 
Then features $\mathcal{X}_1$ and $\hat{\mathcal{X}_2}$ are passed through the two HRNets separately and then fused together via the feature fusion module, prior to the final decision layer. During online inference, our model only requires the first modality (LWIR) as input, and rebuilds the feature of the secondary modality (RGB) conditioned on the first modality. Then the first modality feature and the reconstructed secondary modality feature are passed to the trained model for keypoint estimation inference. This ability to reconstruct features from the unseen modality allows our model to tackle the missing modality issue by training $\hat{\mathcal{X}_2}$ to be an acceptable replacement for $\mathcal{X}_2$.

\begin{table*}[!t]
\caption{Comparative analysis of our approach and baseline methods. Results show the percentages of detections for each joint which are within PCKh@0.5 threshold (in danaLab).}
\vspace{-5pt}
\centering
\label{table:detail-result}
\resizebox{\textwidth}{!}{%
\begin{tabular}{lcccccccccccccccc}
\toprule
Approach                         & Covered & R\_Ankle & R\_Knee & R\_Hip & L\_Hip & L\_Knee & L\_Ankle & R\_Wrist & R\_Elbow & R\_Shoulder & L\_Shoulder & L\_Elbow  & L\_Wrist & Thorax & Head & Total       \\ 
\hline
Fine-tuning                      & ACC     & 95.6 & 96.7 & 86.3 & 90.5 & 95.7 & 95.6 & 90.1 & 93.4 & 92.2 & 94.3 & 93.3 & 88.9 & 98.8 & 98.2 & 93.5            \\
\hline
\multirow{3}{*}{Feature fusion}  & ACC     & 95.7 & 96.9 & 85.0 & 88.9 & 96.0 & 95.0 & 91.6 & 94.9 & 91.3 & 94.5 & 94.2 & 90.8 & 98.9 & 98.3 & 93.7       \\
                                 & OUI       & 98.7 & 97.5 & 94.5 & 94.8 & 97.3 & 98.5 & 95.0 & 95.4 & 94.1 & 96.7 & 95.6 & 94.7 & 99.4 & 99.3 & 96.5       \\
                                 & OCI      & 95.4 & 96.1 & 87.4 & 88.7 & 95.0 & 94.6 & 88.3 & 91.3 & 90.8 & 93.7 & 90.8 & 87.1 & 98.4 & 97.4 & 92.5       \\
\hline
RDF                              & ACC     & 95.5 & 96.6 & 86.7 & 89.0 & 95.7 & 95.0 & 91.0 & 94.2 & 91.2 & 93.8 & 92.6 & 89.3 & 98.7 & 97.8 & 93.4     \\
\hline
Ours                             & ACC     & 96.2 & 96.9 & 88.9 & 91.8 & 95.6 & 95.2 & 91.0 & 94.9 & 93.6 & 95.2 & 93.7 & 89.9 & 99.1 & 98.6 & \textbf{94.3}     \\ 
\bottomrule
\multicolumn{9}{p{350pt}}
{
All cover conditions (ACC), only uncovered images (OUI), only covered images (OCI). \newline
}
\end{tabular}}
\vspace{-14pt}
\end{table*}

\vspace{-8pt}
\section{Experiments and results}

\vspace{-3pt}
\subsection{Dataset and preprocessing}
We train and test all models on the publicly available simultaneously-collected multi-modal lying pose dataset (SLP) \cite{liu2020simultaneously,liu2019seeing}. The SLP dataset has patient pose images taken simultaneously by four different sensors (RGB, LWIR, Pressure mattress and Depth) under three cover conditions: uncovered, lightly covered (bed sheet), and heavily covered (blanket). 
It is collected in two different room environments, namely danaLab and simLab, which have 102 and 7 subjects respectively. Images for danaLab were collected in a home setting, whereas images for simLab were collected in a hospital setting (\textit{e.g.} changes in ceiling heights, hospital beds, types of blankets and participants). 
DanaLab is used during the training and validation process of the developed model. We use the first 70 subjects as the training set, subjects 71-85 form the validation set, and subjects 86-102 form the testing set. All subjects in simLab portion are used to investigate the generalisability of the models.

Since our model has the ability to explicitly learn from the missing modality, to maximize feature extraction and reconstruction of the missing modality, we only consider pairs of LWIR and RGB images which are uncovered, as the covered RGB images are less informative. This will result in the CVAE learning to reconstruct an uncovered RGB image from a covered LWIR image; yet despite the images having different cover conditions they will exhibit the same pose.
We employ the data augmentation scheme used in \cite{liu2020simultaneously}, including rotation, shifting, scaling, color jittering, and synthetic occlusion.

\begin{table}[!t]
\caption{Mean percentages for points which are within PCKh@0.5 threshold, compared with baseline models trained, tested on the corresponding modality and different cover conditions.}
\vspace{-5pt}
\centering
\label{table:ts-result}
\resizebox{0.48\textwidth}{!}{%
\begin{tabular}{lllccc}
\toprule
\multirow{2}{*}{Approach} & Train  & Test & \multirow{2}{*}{Covered} & \multicolumn{2}{c}{PCK@0.5} \\ \cline{5-6} & Modality & Modality & & danaLab & simLab                                      \\ \hline
Sun et al.                & LWIR      & LWIR        & ACC            & 93.5               & 95.3                     \\
\hline
Fine-tuning               & LWIR+RGB         & LWIR     & ACC            & 93.5                 & 95.2         \\
\hline
\multirow{3}{*}{Feature fusion}  & \multirow{3}{*}{LWIR+RGB}  & \multirow{3}{*}{LWIR+RGB}  & ACC  & 93.7 & 95.6       \\
                                 &                            &                            & OUI    & 96.5 & 95.8       \\
                                 &                            &                            & OCI   & 92.5 & 93.7       \\
\hline
RDF                              & LWIR+RGB                   & LWIR                       & ACC  & 93.4 & 94.6      \\
\hline
Ours                             & LWIR+RGB                   & LWIR                       & ACC  & 94.3 & \textbf{95.8}     \\ 
\bottomrule
\multicolumn{6}{p{320pt}}
{
All cover conditions (ACC), only uncovered images (OUI), only covered images (OCI). \newline
}
\end{tabular}}
\vspace{-18pt}
\end{table}

\vspace{-3pt}
\subsection{Experimental setup and evaluation metric}

We train and evaluate our fine-tuning and RDF baseline on LWIR and RGB modalities with all three cover conditions combined. For our feature fusion baseline model, we feed it with both modalities, but to investigate the effect of different cover conditions, we separate cover conditions into three combinations: i) all cover conditions (ACC), ii) only uncovered images (OUI), and iii) only covered images (OCI).

In this evaluation, we use the PCKh with a threshold of $0.5$ as our evaluation metric. For a fair comparison, we employ HRNet\cite{sun2019deep} as our backbone for all models, all models are trained with the same setting as \cite{liu2020simultaneously}. 
All models are trained from scratch, for 100 epochs, with a learning rate 1e-3 and the Adam optimizer~\cite{kingma2014adam} (learning decay rate 0.1 at epoch 70 and 90). All models have the same HRNet setting with the W32 configuration~\cite{sun2019deep}. All models which involve feature fusion use the same fusion method, IAFF~\cite{dai2021attentional}.
To demonstrate the effectiveness of our feature fusion and reconstruction method, we calculate the mean of the importance weight kernel $\mathbf{W}$ of
\begin{equation}
    \mathcal{F}_{\mathrm{fused}}=(1-\mathbf{W})*\mathcal{F}_{\mathrm{LWIR}} + \mathbf{W}*\mathcal{F}_{\mathrm{RGB}} 
\end{equation}
in the IAFF module \cite{dai2021attentional}. This is a sound indicator of how much our models utilize features from the RGB channel.

\begin{figure}[!t]
\centering
\includegraphics[width=0.86\linewidth]{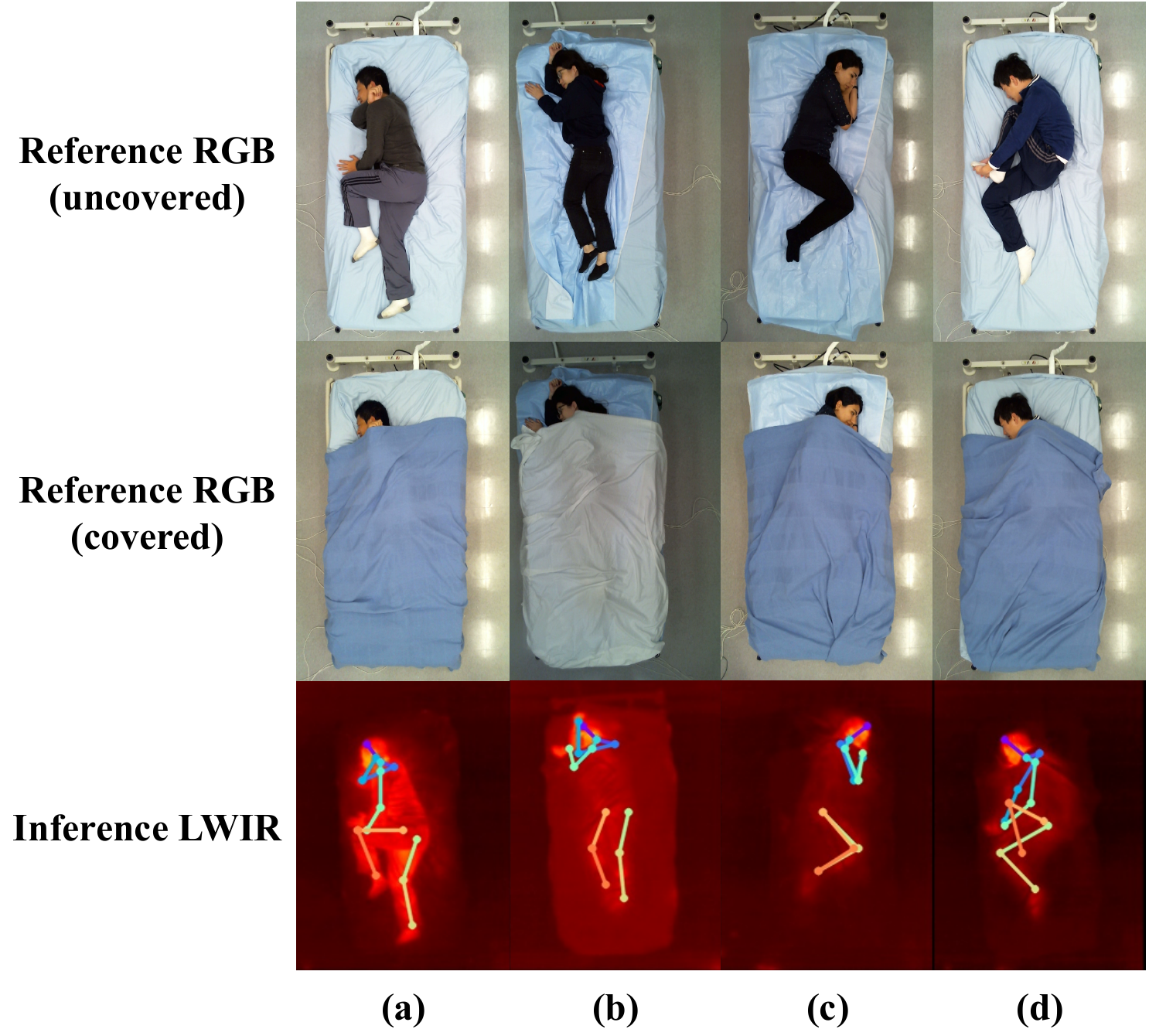}
\vspace{-3pt}
\caption{
Qualitative results of human pose estimation on covered and uncovered LWIR images. RGB images are only shown to aid results for visualisation. Inference from (a) uncovered, (b-d) covered.
}
\label{fig:Fig2}
\vspace{-13pt}
\end{figure}

\vspace{-3pt}
\subsection{Results and discussion}
As shown in Table~\ref{table:detail-result} and Table~\ref{table:ts-result}, under the same constraints (same cover condition, same inference domain), our proposed model outperforms the RDF and Fine-tuning methods, and achieves 94.3\% accuracy for PCKh@0.5 on the danaLab test set, and 95.8\% accuracy on the simLab test set. Qualitative results are presented in Fig.~\ref{fig:Fig2}.
%
%
The RGB domain carries rich visual information such as color and depth, that can provide more clues for the pose estimation models for uncovered images. Further, using this information leads to better performance for the feature fusion based model in the presence of uncovered images, as shown in Table~\ref{table:detail-result}. Therefore, reconstructing these features through our proposed model can be beneficial and increases the performance of the pose estimation, even in the absence of that modality as shown in Table~\ref{table:ts-result}.

Furthermore, during inference, our model fuses the features from the LWIR modality and the ones reconstructed from the RGB modality with an average ratio of 55:45. This indicates that our model can fully leverage the reconstructed RGB features during inference, which is a noticeable advantage compared to the RDF model. 

In most hospitals, capturing RGB images are prohibited due to preserving patients' privacy. Further, RGB images include highly occluded scenarios with bedding which can hamper any pose estimation model. Our baseline models are limited by those conditions. Feature fusion based models cannot work on a single modality input, and their performance can be negatively impacted if patients are covered. RDF will not use RGB features during inference, as it sets all (unavailable) RGB images to $0$. In contrast, our model continues to work under these two limitations and does not require the presence of both modalities during inference. Furthermore, it can learn to reconstruct uncovered RGB features from covered LWIR features, and effectively use these during inference.

\vspace{-8pt}
\section{Conclusion}
\vspace{-5pt}

In this paper, we propose a novel approach for incorporating self-supervised feature extraction into state-of-the-art human pose estimation approaches (e.i HRNet) to reconstruct a missing modality and provide accurate in-bed pose estimation when only privacy preserving imaging modes are available for the AI model deployment.
Our findings in this study demonstrate a promising deep learning-based solution that can improve inpatient monitoring in the presence of scarce annotations, which has historically imposed severe limitations on machine learning methods in the medical domain.

\section{Compliance with Ethical Standards}
\vspace{-5pt}
This research study was conducted retrospectively using human subject data made available in open access by Ostadabbas et al.~\cite{liu2020simultaneously,liu2019seeing}. Ethical approval was not required as confirmed by the license attached with the open access data.
The experimental procedures involving human subjects data described in this paper were approved by the CSIRO Health and Medical Human Research Ethics Committee (CHMHREC). 





\bibliographystyle{IEEEbib}
\bibliography{refs}



 


\end{document}